\documentclass[preprint,12p]{elsarticle}

\usepackage{amssymb}

\usepackage{amsmath,amsfonts}
\usepackage{comment}
\usepackage{subfigure}
\usepackage{algorithmic}
\usepackage{array}
\usepackage{caption}
\usepackage{textcomp}
\usepackage{stfloats}
\usepackage{url}
\usepackage{verbatim}
\newcolumntype{Y}{>{\centering\arraybackslash}X}
\usepackage{tikz}
\usepackage{enumerate} 
\usepackage{tabularx}
\usepackage [english]{babel}
\usepackage [autostyle, english = american]{csquotes}
\MakeOuterQuote{"}
\usepackage{amssymb,amsmath, amsthm, amstext, graphics, color}
\usepackage{todonotes}
\usepackage{graphicx}
\graphicspath{ {images/} }
\usepackage{multicol}
\usepackage{color}
\usepackage{enumitem}
\setlist{nolistsep}
\usepackage{hyperref}
\usepackage{comment}
\usepackage{pgf,tikz}
\usepackage{calrsfs}
\usetikzlibrary{arrows,shapes}

\usepackage{algorithmic}
\usepackage{algorithm}
\usepackage{pgfplots}

\usepackage{balance}
\usepackage{graphicx}

\pgfplotsset{compat=1.5.1}

\def\BibTeX{{\rm B\kern-.05em{\sc i\kern-.025em b}\kern-.08em
    T\kern-.1667em\lower.7ex\hbox{E}\kern-.125emX}}

\begin{document}

\begin{frontmatter}



\title{Concise Fuzzy Planar Embedding of Graphs: a Dimensionality Reduction Approach}


 \affiliation[label1]{organization={ Department of Computer Science and Mathematics, Lebanese American University},city={Beirut},country={Lebanon}}
             

 \author[label1]{Faisal N. Abu-Khzam\corref{cor1}}
 \author[label1]{Rana H. Mouawi}
 \author[label1]{Amer Hajj Ahmad}
 \author[label1]{Sergio Thoumi}
 
\cortext[cor1]{Corresponding author, \textit{email address:} faisal.abukhzam@lau.edu.lb}

\begin{abstract}
The enormous amount of data to be represented using large graphs exceeds in some cases the resources of a conventional computer. Edges in particular can take up a considerable amount of memory as compared to the number of nodes. However, rigorous edge storage might not always be essential to be able to draw the needed conclusions. A similar problem takes records with many variables and attempts to extract the most discernible features. It is said that the ``dimension'' of this data is reduced. Following an approach with the same objective in mind, we can map a graph representation to a $k$-dimensional space and answer queries of neighboring nodes mainly by measuring Euclidean distances. The accuracy of our answers would decrease but would be compensated for by fuzzy logic which gives an idea about the likelihood of error. 
This method allows for reasonable representation in memory while maintaining a fair amount of useful information, and allows for concise embedding in $k$-dimensional Euclidean space as well as solving some problems without having to decompress the graph. Of particular interest is the case where $k=2$.
Promising highly accurate experimental results are obtained and reported.

\end{abstract}



\begin{keyword}
Graph compression \sep graph embedding \sep fuzzy logic \sep fastmap \sep dimensionality reduction.

\end{keyword}

\end{frontmatter}


\section{Introduction}

The constant growth of social networks and the corresponding immense size of graph representation impose some difficulties in maintaining and manipulating such data. In particular, the edges that make up a graph can exceed its corresponding nodes by a great margin so that classical representation using an adjacency matrix/list is nearly impossible when dealing with big graphs. This is especially true when using a system with a conventional main memory. Other notable examples of big graphs include graphs that are used to represent webpages and their interconnections (i.e., hyperlinks) which can have close to 20 billion nodes and 160 billion edges. Social networks, on the other hand, can contain a billion users with relationships among users reaching more than 140 billion. 

The amount of memory needed to work on such graphs gave rise to the need of "graph compression" or "compact graph representation." Similar to the compression of data, such as images and videos, graph compression can be lossy or lossless. In lossy compression, some information about the graph's structure are lost, so for example two adjacent nodes might not be adjacent in the compressed version of the graph as some edges are not present anymore. However, in some cases when extracting useful information from graphs, it is possible that not all edges are of the utmost importance to reach a conclusion. That is, some inaccuracy might be tolerated during analysis. This is especially the case if we are able to get an estimate of whether or not two nodes are neighbors. The result is that we find a compromise between maintaining all data in a graph and being able to store that graph in a typical main memory. All the nodes in an input graph are preserved while providing an alternate representation of the adjacency matrix. 

The need for concise and accurate graph representation is also due to a large number of applications in neural networks, which obviously require their input data to be in vector-form \cite{Dai16,Dai18,abuObs,Bello2016, NIPS2015_5866, Zhang2018, Gori2005ANM,FrasconiAdaptiveProcessing,ScarselliGNN}. 
Some data types can be easily mapped (or embedded) into vectors, but graphs pose their own challenge. An obvious choice would be to use a graph's adjacency matrix since it is already in a desired format. However, this can be inefficient for large graphs since they typically have sparse adjacency matrices \cite{sanchez2021gentle}. Many graph embedding methods have been proposed in the literature, mostly based on unsupervised learning  \cite{ge1,ge2,ge3,ge4,ge5}. However, most existing methods incur  notable information loss and accuracy of graph embedding remains a challenge. 

In this paper, we present a new graph compression approach that moves us from needing a quadratic amount of storage in the number of nodes to a linear amount while preserving the adjacency relation as much as possible. The main idea is to simply map the nodes of a graph into points in a two-dimensional Euclidean space so that adjacency is modeled via proximity. The nodes will be represented by the coordinates of their corresponding points so that two nodes are definitely adjacent if they are ``close enough'' and definitely non-adjacent if they are ``far enough.'' On the other hand, nodes whose corresponding points are neither close nor far will be judged upon via fuzzy logic. The proposed approach  presents an ideal method for graph embedding as it represents graphs in a $n \times 2$ matrix, which makes it more appealing for neural networks. Unlike adjacency matrices, and in addition to nearly-perfect adjacency prediction, our resulting matrix would be used entirely.



This paper is structured as follows. In the next section we provide a brief overview of the ``graph compression'' literature. Then, in Section 3, we present our fuzzy graph mapping approach. Section 4 is devoted to experimental results, and we discuss graph compression methods as well as some related research topics. In Section 5, we present more results and some future directions. 
We conclude in Section 6.   

\section{Literature Review}

Currently, several frameworks exist to deal with sizeable graphs such as Pregel \cite{Pregel}, Apache Giraph \cite{giraph}, SweG \cite{sweg}, and GraphLab \cite{graphlab}. These projects are based on leveraging the computing resources of many machines so that the operations on the large graphs are distributed. On the other hand, one might opt for finding a more packed representation of the graph. A recent approach for graph compression relies on identifying repeated patterns in graphs and representing them through ``grammar rules''\cite{MP17}. This latter approach aims to enhance the performance of certain types of queries. In addition to these, there are multiple known approaches that will be presented in this section.

The recent literature includes a notable number of lossy graph compression algorithms.
They can differ in the method used, objectives, and results achieved (compression ratio/accuracy). For applications where certain graph properties need to be preserved, the Slim graph framework \cite{oltchik} can be used, it preserves certain properties such as Minimum Spanning Trees (MSTs) and connected components.
Another algorithm that can be used for this purpose is Graph Compression With Embedding (G-CREWE) \cite{gcrewe}, which preserves important topological structures. G-CREWE uses a Graph Convolutional Network (GCN) for feature extraction and learning node embedding, then this is used to help with their lossy compression.

Many current algorithms do not deal with weighted graphs. However, this might still be needed for some applications such as cost minimization. Toivonen et al. \cite{toivonen} presented a method for the compression of weighted graphs. They achieved a compression ratio up to two with an error factor between $0.03$ and $0.06$ for every node pair. 

In some applications, dynamic graphs that are constantly updated are of interest. This poses a new challenge to the standard compression techniques. MoSSo \cite{komosso}, a lossless algorithm, can be used for the compact representation of dynamic graphs. The algorithm is incremental and it works by moving nodes among supernodes in order to represent the changes happening in the graph.
Another dynamic compression algorithm is StarZIP \cite{8649624} that deals with streaming graphs. It divides a graph into a uniform representation of stars to achieve up to 60X compression.
 
Designing algorithms for WebGraphs is still of interest. Some of the notable recent algorithms in this area include SSumM \cite{ssumm}, which is a lossy graph compression algorithm that is scalable. SSumM works by merging nodes together and sparsifying the compressed graph. The summary graphs obtained are up to 11.2X smaller while having a small reconstruction error. Finally, Apostolico and Drovandi \cite{a2031031} presented a graph compression algorithm based on Breadth First Search (BFS) traversal, it achieved a compression of up to 5 bits per link. 

So far, most of the graph compression methods in the literature bear something in common. They mainly consist of merging multiple nodes in a graph into ``supernodes'' while multiple edges are merged into superedges. Another common  practice is the use of distributed algorithms in order to handle large graphs, assuming they do not fit in the RAM of a single machine.
In this paper, we propose a novel method that is based on dimensionality reduction. The main idea stems from the fact that every graph can be represented as a geometric intersection graph of $d$-dimensional disks (or spheres/balls). To explicate, an intersection graph of $d$-dimensional disks (henceforth $d$-balls) is a simple undirected unweighted graph with nodes represented by $d$-balls; two nodes are adjacent (i.e., linked by an edge) if the corresponding balls intersect. 

Transforming a graph into the intersection graph of $d$-balls of the same radius in $NP$-hard \cite{BREU19983,KangM12}.
Moreover, a mere transformation into such an intersection graph can result in a high-dimensional representation. 
To cope with these difficulties, our approach consists of mapping each node of a graph into two (corresponding) concentric $d$-balls. The points that lie within the smaller ball will correspond to what the node can see as neighbors (or out-neighbors in directed graphs) while points that lie outside the larger ball correspond to nodes that are too far to be (out)neighbors. Consequently there would be some ``uncertainty range'' (between the two $d$-balls), for which we use a fuzzy system to judge whether two nodes are adjacent or not. 

To achieve our objective, the first phase of our algorithm consists of mapping the nodes of a given graph into a two-dimensional Euclidean space, using a dimensionality reduction algorithm. The approach could be used with any $k-$ dimensional space. The most notable dimensionality reduction methods in the literature are Multidimensional Scaling (MDS) \cite{mds} and Principal Component Analysis (PCA) \cite{wold1987principal}.
Both methods are based on a linear transformations of high-dimensional data elements (or points) into a lower-dimensional space.
While MDS and PCA can be optimal in the sense that they can reduce the data representation into (what is known as) its intrinsic dimension, their running time is at best quadratic in the number of data elements, which makes these methods prohibitively inefficient on large data sets. FastMap was introduced as an efficient and effective alternative \cite{faloutsos95fastmap}. It was shown to preserve how data elements cluster \cite{Khan}. 
The Fastmap algorithm takes a distance matrix and returns a set of points in a $k$-dimensional space, where $k$ is user defined.
Unlike MDS, Fastmap yields an approximate dimensionality reduction. Yet, it was shown to be effective when it comes to the resulting {\em stress function} \cite{faloutsos95fastmap}, which is nothing but the average error of the new distances (in the FastMap mapping) relative to the distances in the input.   
Fastmap runs in linear-time in the number of data elements (graph nodes in our case), thus it outperforms the other algorithms.

\section{A Graph Mapping Approach}

In order to manage the large amount of data associated with graphs, we seek to transform
their representation to a more compact form, even if some minimal amount of information is lost. 
The graphs we consider can be either directed or undirected. We will describe our approach for undirected graphs first, then we show how the same method is applied to directed graphs. 

As mentioned earlier, our proposed method is to represent the nodes as points, or vectors, in a two-dimensional space, with distances between them indicating their adjacency status. To answer adjacency queries, each node $v$ is associated with two parameters, $r(v)$ and $R(v)$, that indicate the ``radius'' of $d$-ball within which the points corresponding to its neighbors are located and the ``radius'' outside of which its non-neighbors' points are located, respectively. There is uncertainty regarding the points that lie within these two values.

For the first part, we use the linear-time FastMap algorithm from \cite{faloutsos95fastmap} after transforming the given graph's adjacency matrix into a distance matrix. This is necessary because Fastmap takes a distance matrix as input and returns a set of points in a $k$-dimensional space where $k$ is to be selected based on the desired accuracy, and it can also be user defined.

Once we have the above described mapping, 
it is possible to calculate the two parameters $r$ and $R$ of each node $v$. This is done by calculating the Euclidean distance from $v$ to all other nodes, and arranging those distances, for example, in an increasing manner. 


Iterating from the beginning of this list, we determine the least ``$r = r(v)$'' such that all nodes within it are neighbors of $v$. This guarantees that any node whose corresponding point is within distance $r(v)$ from $v$ is a neighbor of $v$ 
Similarly, from the end of the list we determine the value of ``$R=R(v)$.'' 
Of course, the resulting representation will not give a definitive ``yes'' or ``no'' about the adjacency of two nodes unless it is truly the case.


When the Euclidean distance $d$ between the two nodes $v$ and $v'$ lies within the values $r$ and
$R$ of both $v$ and $v'$, we invoke a fuzzy logic system that gives us the likelihood of those two nodes being neighbors, with larger outputs indicating a higher chance of them being adjacent. 
This system assumes the closer $d$ is to the value of $r(v)$ as compared to $R(v)$, 
the more likely it is that $v$ and $v'$ are adjacent. The opposite is assumed when the measured distance is closer to the value of $R(v)$. 
Furthermore, we compute for each node $v$ the number of neighboring nodes between the two regions, or balls, of radii $r(v)$ and $R(v)$ divided by the number of all nodes in that region. This latter value is used as an additional heuristic to increase the fuzzy accuracy as the graph gets denser.
We pre-process two arrays when calculating the $r$ and $R$ values for each node: $neighborsInsider$ and $nodesInBetween$, where $neighnorsInsider[v]$ is the number of neighbors that are inside $r(v)$, and $nodesInBetween[v]$ is the number of nodes that are between $r(v)$ and $R(v)$.
Then, we calculate the second heuristic $inRegionDegree[v]$ for a node $v$ as:

\begin{equation*}
\begin{split}
\frac{degree[v] - neighborsInsider[v]}{nodesInBetween[v]}
\end{split}    
\end{equation*}

The system takes as input two crisp values between 0 and 1. The first value obtained by dividing the difference of $R$ and $d$ by the difference of $R$ and $r$. Whereas, the second value is obtained by dividing the number of neighbors between $r$ and $R$ by the number of overall nodes in this region. The inputs are subjected to three membership functions that
correspond to the likelihood of the queried nodes to be neighbors. The inference system relies on two (successive) rules: (1) if the first input value $EstimatedDistance$ is ``close to smaller $r$'' AND the second input value $InRegionDegree$ is ``High node degree,'' then the two nodes are neighbors. If this is not the case,  and (2) if the first input value is ``close to larger $R$'' AND the second input value is ``Low node degree,'' then the two nodes are not neighbors. 



\begin{figure}[!htb]
\begin{tikzpicture}
\begin{axis}[
    legend style={nodes={scale=0.8, transform shape}},
    axis lines = left,
    xmin=0, xmax=1,
    ymin=0, ymax=1,
    xtick={0,0.2,0.4,0.6,0.8,1},
    ytick={0,0.2,0.4,0.6,0.8,1},
    legend pos=south west,
    ymajorgrids=true,
    grid style=dashed,
]
\addplot [
    domain=0:1, 
    samples=10, 
    color=red,
]
    {-1.428*x + 1};
\addlegendentry{Close to larger R}
\addplot [
    domain=0:1, 
    samples=10, 
    color=blue,
    ]
    {2.5*x - 1.5};
\addlegendentry{Close to smaller r}
 
\end{axis}
\end{tikzpicture}
\centering
\vspace{-5pt}
\caption{First input membership function (estimated distance)}
\end{figure}

\begin{figure}[!htb]
\begin{tikzpicture}
\begin{axis}[
    legend style={nodes={scale=0.8, transform shape}},
    axis lines = left,
    xmin=0, xmax=1,
    ymin=0, ymax=1,
    xtick={0,0.2,0.4,0.6,0.8,1},
    ytick={0,0.2,0.4,0.6,0.8,1},
    ymajorgrids=true,
    grid style=dashed,
]
\addplot [
    domain=0:1, 
    samples=100, 
    color=red,
]
    {1 / ( 1 + exp( 4 * ( x - 0.65 ) ) )};
\addlegendentry{Low node degree}
\addplot [
    domain=0:1, 
    samples=100, 
    color=blue,
    ]
    {1 / ( 1 + exp( -4 * ( x - 0.65 ) ) )};
\addlegendentry{High node degree}
 
\end{axis}
\end{tikzpicture}
\centering
\vspace{-5pt}
    \caption{Second input membership function (in region degree)}
\end{figure}

Note that the AND operator in the fuzzy logic inference system corresponds to taking the minimum value. These rules are specified in an ``fcl'' file in a fuzzy control
language \cite{CingolaniA12}. There are also three
membership functions for the output set that correspond to whether two veritces are adjacent or
non-adjacent. The activator of the
inference rules is the minimum operator which truncates the output membership functions for each rule. The functions of the input and output sets are represented in Figures 1-3. 
Accumulation of the inference rules' results takes place using the
maximum operator. Finally, defuzzifying the output value is done using the center of gravity method \cite{van1999defuzzification}.

\begin{figure}[!htb]
\begin{tikzpicture}
\begin{axis}[
    legend style={nodes={scale=0.8, transform shape}},
    axis lines = left,
    xmin=0, xmax=1,
    ymin=0, ymax=1,
    xtick={0,0.2,0.4,0.6,0.8,1},
    ytick={0,0.2,0.4,0.6,0.8,1},
    ymajorgrids=true,
    grid style=dashed,
]
\addplot [
    domain=0:1, 
    samples=10, 
    color=red,
]
    {-2*x + 1};
\addlegendentry{Not Neighbors}
\addplot [
    domain=0:1, 
    samples=10, 
    color=blue,
    ]
    {2*x - 1};
\addlegendentry{Neighbors}
 
\end{axis}
\end{tikzpicture}
\centering
\vspace{-5pt}
    \caption{Output membership function}
\end{figure}


An example of our approach is presented in Figures \ref{sample-input-graph} and \ref{2D-example}.
Each node will be mapped to a point in a 2-dimensional space using FastMap, yielding the following alternate representation. The values for $r$ and $R$ are also calculated for each node. 
The two circles that correspond to node 2 are shown in Figure \ref{2D-example}. 
Nodes 1 and 5 lie within the smaller circle, which indicates they are definitely neighbors of node 2.
Node 6 lies strictly outside the larger circle, so it is definitely a non-neighbor. The fuzzy inference system will report that 3 is a neighbor, being closer to the smaller circle while node 4 is a non-neighbor because it lies on the outer circle.

\begin{figure}[htb!]
\begin{tikzpicture}
  [scale=.7,auto=left,every node/.style={circle,fill=blue!20}]
  \node (n6) at (3.5,6) {6};
  \node (n4) at (5,8)  {4};
  \node (n5) at (8,8)  {5};
  \node (n1) at (11,8) {1};
  \node (n2) at (9.5,10)  {2};
  \node (n3) at (6.5,10)  {3};

  \foreach \from/\to in {n6/n4,n4/n5,n5/n1,n1/n2,n2/n5,n2/n3,n3/n4}
    \draw (\from) -- (\to);

\end{tikzpicture}
\centering
\caption{A sample input graph.}
\label{sample-input-graph}
\end{figure}

\begin{figure}[!htb]
\includegraphics[scale=1.7]{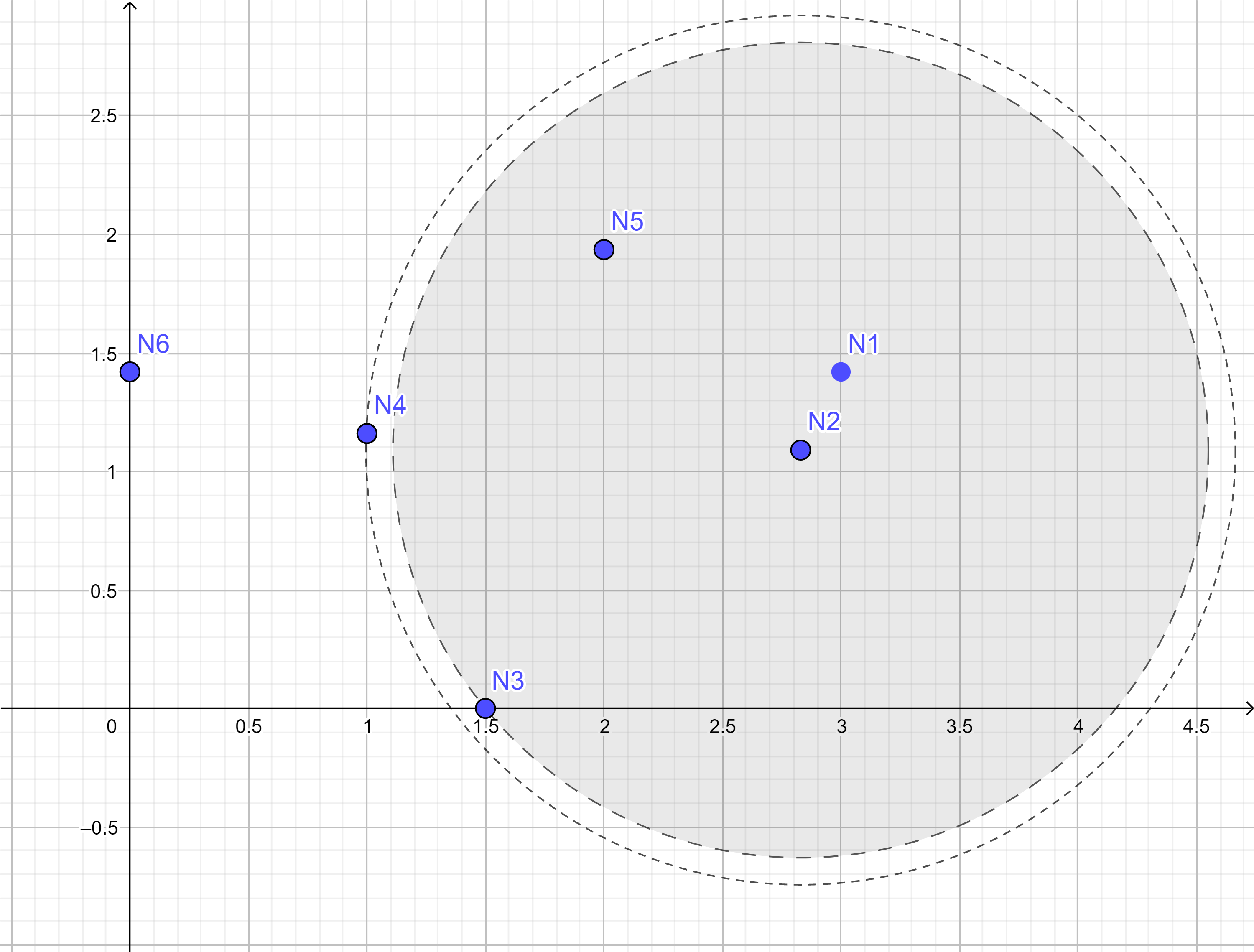}
\centering
\caption{Mapping of graph nodes to 2-dimensional space}
\label{2D-example}
\end{figure}

The above method is used in the below pseudocode for the adjacency query algorithm, which assumes the two vectors (or arrays) $r$ and $R$ have been computed as described above. The algorithm uses the output of the Fastmap algorithm to determine whether two nodes $v$ and $v'$ are
adjacent. The value of $d$ is nothing but the Euclidean distance between the points that correspond to $v$ and $v'$.

\begin{algorithm}[!htb]
\caption{Adjacency Query - Undirected Graphs}
\begin{algorithmic}[1]
	\IF{d $\leq$ $r(v)$ \OR d $\leq$ $r(v')$}
		\RETURN 1
    \ELSIF{d $\geq$ $R(v)$ \OR d $\geq$ $R(v')$}
    	\RETURN 0
    \ENDIF
    
    \STATE $EstimatedDistance1$ = $(R(v) - d)/(R(v) - r(v))$
    \vspace{5pt}
    \STATE $InRegionDegree1$ = $\frac{NeighborsInRegion(v)}{NodesInRegion(v)}$
    \vspace{5pt}
    \STATE $output1$ = FuzzyLogicSystem($EstimatedDistance1$, $InRegionDegree1$)
    \STATE $EstimatedDistance2$ = $(R(v') - d)/(R(v') - r(v'))$
    \vspace{5pt}
    \STATE $InRegionDegree2$ = $\frac{NeighborsInRegion(v')}{NodesInRegion(v')}$
    \vspace{5pt}
    \STATE $output2$ = FuzzyLogicSystem($EstimatedDistance2$, $InRegionDegree2$)
    \RETURN minimum($output1$, $output2$)
\end{algorithmic}
\end{algorithm}

We note that our approach works for directed graphs as well. 
The query in this case takes an ordered pair $(v,v')$ as input and the condition for a yes-answer would simply depend on $r(v)$ and $R(v)$ only. In fact, the condition $d \leq r(v)$ would be enough to conclude that there is an arc (or directed edge) from $v$ to $v'$. Moreover, in the case of uncertainty, we would only compute the values of $EstimatedDistance1$ and $InRegionDegree1$, then $output1$ would be returned. This is illustrated in the below pseudocode.

\begin{algorithm}[!htb]
\caption{Adjacency Query - Directed Graphs}
\begin{algorithmic}[1]
	\IF{d $\leq$ $r(v)$}
		\RETURN 1
    \ELSIF{d $\geq$ $R(v)$}
    	\RETURN 0
    \ENDIF
    
    \STATE $EstimatedDistance$ = $(R(v) - d)/(R(v) - r(v))$
    \vspace{5pt}
    \STATE $InRegionDegree$ = $\frac{NeighborsInRegion(v)}{NodesInRegion(v)}$
    \vspace{5pt}
    \STATE $output$ = FuzzyLogicSystem($EstimatedDistance$, $InRegionDegree$)
    \RETURN $output$
\end{algorithmic}
\end{algorithm}


Our graph mapping approach can be split into two different methods based on how we determine the distance between any two nodes in the given graph. 
Since our graphs are unweighted, computing all the pair-wise distances between two nodes can be performed via $n$ calls to Breadth-First Search. The overall running time would be in  $\mathcal{O}(n^2+ne)$ where $n$ and $e$ are the numbers of nodes and edges, respectively. This can be prohibitively slow on large graphs. To cope with this limitation, we can apply a depth-limited BFS: run BFS from each node to cover nodes within distance $t$, for some user-defined parameter $t$. The distance between two nodes would be either $\leq t$ (if they are at most $t$-steps away from each other) or $n$. When $t=1$, the distance matrix is simply obtained from the adjacency matrix by replacing each 0 by $n$ except for diagonal cells. The smaller the value of $t$, the more efficient this very first step is. In our experiments, and since we use a single machine and deal with large graphs, we mainly used $t=1$. We refer to it as the binary-distances' method since any two nodes are either at distance 1 or $n$.
This has reduced the running time from hours to minutes on some large graphs. Obviously, the gain in performance comes with a (very small) reduction in accuracy.


\section{Experiments}


Experiments were conducted on a number of known data sets, namely: 
the ArXiv Collaboration graphs \cite{leskovec2007graph}; email-Enron \cite{klimt2004introducing}; Brightkite networks \cite{chofriendship} from the Stanford Network Analysis Project (SNAP) \cite{snapnets}; as well as the Catster households dataset from the KONECT project \cite{kunegis2013konect}. 
The first table below shows the properties of each dataset in terms of number of nodes and edges per graph.

\vspace{10pt}

\begin{table}[!htb]
\centering
\begin{tabularx}{\columnwidth}{|Y|Y|Y|}
\hline
{\bf Dataset Name} & {\bf Nodes} & {\bf Edges}  \\ \hline
ca-AstroPh  & 18772 & 198110 \\ \hline
ca-CondMat   & 23133 & 93497  \\ \hline
ca-GrQc      & 5242  & 14496  \\ \hline
ca-HepPh     & 12008 & 118521 \\ \hline
ca-HepTh     & 9877  & 25998  \\ \hline
Brightkite & 58228  & 214078 \\ \hline
email-Enron &  36692  &  183831 \\ \hline
Catster households & 	105138 & 	494858 \\ \hline 
\end{tabularx}
\caption{Data sets used.}
\label{tab:datasets}
\end{table}

The very first step consists of building a distance matrix from the given adjacency matrix, or list. In our experiments, and since we deal with large graphs, we have adopted the binary-distances method discussed earlier, simply because of its efficiency compared to exact distances (seconds versus days). 
Moreover, the binary distances method exhibited the same (close to 100\%) fuzzy accuracy as the exact distances' method.
Nevertheless, and for the sake of comparison, we compute the exact distances for modest-size graphs, such as the ArXiv Collaboration graphs. In particular, if our fuzzy system is not to be used, we show that exact distances can help us obtain a more accurate graph compression (than binary distances) via a mere dimensionality reduction approach. For this purpose, and after applying Fastmap in the first (graph mapping) phase, we can check whether the value of $k$ is close or equal to the graph's intrinsic dimension simply by checking the percentage of definite answers. This is the ratio of correctly answered adjacency queries to all possible queries (i.e., covering all pairs of nodes).
The below diagram shows the percentages of definite answers on the ca-GrQc graph from the ArXiv Collaboration dataset, obtained after using binary distances and exact distances, respectively.

\vspace{5pt}

\begin{figure}[htb!]
\centering
\begin{tikzpicture}
\begin{axis}[
    ybar,
    width=\columnwidth,
    height=8cm,
    ymin=0,
    ymax=100,
    ylabel={Percentage of Definitive Answers},
    xlabel={$k$},
    xtick=data,
    xticklabels={2, 50, 100, 200, 300, 400},
    ytick={0, 10, 20, 30, 40, 50, 60, 70, 80, 90, 100},
    legend style={at={(0.5,1.15)},
      anchor=north,legend columns=-1},
    symbolic x coords={2, 50, 100, 200, 300, 400},
    bar width=15pt,
]

\addplot[fill=blue] coordinates {
    (2, 56.43)
    (50, 58.65)
    (100, 60)
    (200, 82.18)
    (300, 92.21)
    (400, 99.81)
};
\addplot[fill=red] coordinates {
    (2, 0.114)
    (50, 15.703)
    (100, 25.501)
    (200, 37.421)
    (300, 45.32)
    (400, 55.96)
};

\legend{Exact, Binary}

\end{axis}
\end{tikzpicture}
\caption{Definitive answers percentage on the ca-GrQc dataset}
\label{definite}
\end{figure}

We now present our main results for lossy graph compression. The accuracy of fuzzy adjacency queries and the graph  compression ratio are used as quality measures. The compression ratio is widely defined as:

\begin{equation*}
\begin{split}
\frac{uncompressed ~size}{compressed ~size}   
\end{split}    
\end{equation*}

In our case, this is calculated for undirected graphs represented by an adjacency matrix as: 

\begin{equation*}
\begin{split}
\frac{n^2}{nk} = \frac{n}{k}
\end{split}    
\end{equation*}



Our experiments show that the fuzzy accuracy was surprisingly very high and (therefore) varied insignificantly with the choice of $k$. As such, the choice of $k=2$ would be recommended for our fuzzy graph representation since it obviously gives the highest compression ratio. 

For an adjacency list representation, and with $k=2$, our dimensionality reduction method still achieves notable compression ratios. For example, if the number of edges is only three times the number of nodes ($e=3n$), then the compression ratio would be two. To see this, note that the uncompressed graph size would be $n+e=4n$. 
As far as we know, the number of edges is usually larger (than $3n$) in all large graphs used in the literature. Obviously, there is a positive relationship between the number of edges and the compression ratio in the case of an adjacency list representation.

The results of our experiments on the graphs of Table \ref{tab:datasets} are presented in Table \ref{tab:merge1} below. Obviously, the achieved fuzzy accuracy was nearly perfect and resulted in a nearly perfect overall accuracy, thus achieving nearly lossless compression.
A fuzzy answer is considered ``sound'' if it returns a value above 0.5 for a pair of nodes that happen to be neighbors, and a value below 0.5 for a pair that are non-neighbors. The graph with the red line in Figure \ref{Accuracy} shows, among fuzzy queries, the percentage of sound ``yes'' and sound ``no'' answers as a function of the target dimension $k$.
While the definite answers' results of Figure \ref{definite} suggest a tremendous loss of information for small values of $k$, our results show that fuzzy answers can almost always make up for more certain ones. 

\begin{table}[!htb]
\centering
\begin{tabularx}{\columnwidth}{|Y|Y|Y|Y|}
\hline
{\bf Dataset} & {\bf Fuzzy Accuracy} & {\bf Overall Accuracy} & {\bf Compression Ratio} \\ \hline
ca-GrQc & 99.99        & 99.89        & 2621              \\ \hline
ca-AstroPh & 99.99      & 99.88         & 9386                 \\ \hline
ca-HepPh & 99.99       & 99.83           & 6004                \\ \hline
ca-HepTh & 99.99        & 99.94          & 4938.5              \\ \hline
ca-CondMat & 99.97      & 99.96           & 11566.5                    \\ \hline
Brightkite &    100          &          99.99        &   29114                          \\ \hline
email-Enron &     100        &     99.97             &     18346                      \\ \hline
Caster households &    \vspace{0.5pt} 100   & \vspace{0.5pt}    99.98              &    \vspace{0.5pt}      34428                      \\ \hline
\end{tabularx}
\caption{Results with the binary method for $k=2$}
\label{tab:merge1}
\end{table}

\begin{figure}[!htb]
\begin{tikzpicture}
\begin{axis}[
    legend style={nodes={scale=0.8, transform shape}},
    xlabel={$k$ (number of dimensions)},
    ylabel={Accuracy},
    xmin=0, xmax=100,
    ymin=0, ymax=110,
    xtick={0,20,40,60,80,100},
    ytick={0,20,40,50,60,80,90,100},
    legend pos=south east,
    ymajorgrids=true,
    grid style=dashed,
]
 
\addplot[
    color=red,
    mark=dot,
    ]
    coordinates {
    (2,96.91)(3,96.67)(5,96.41)(8,93.6)(10,93.298)(12,93.94)(17,94.31)(23,94.35)(30,94.48)(38,94.46)(47,94.44)(57,94.41)(68,94.45)(80,94.31)(93,94.51)
    };
    \addlegendentry{$Fuzzy Accuracy$}
    
\addplot[
    color=blue,
    mark=dot,
    ]
    coordinates {
     (2,7.349)(3,13.66)(5,18.09)(8,43.34)(10,46.3)(12,44.64)(17,47.55)(23,48.8)(30,48.18)(38,47.88)(47,48.02)(57,48.42)(68,48.166)(80,49.2)(93,46.89)
    };
    \addlegendentry{$Definitive Answers Accuracy$}
    
\addplot[
    color=black,
    mark=dot,
    ]
    coordinates {
    (2,97.13)(3,97.128)(5,97.06)(8,96.36)(10,96.39)(12,96.64)(12,97.08)(17,97)(23,97.1)(30,97.14)(47,97.11)(57,97.11)(93,97.0875)
    };
    \addlegendentry{$Overall Accuracy$}

\end{axis}
\end{tikzpicture}
    \centering
    \vspace{-10pt}
    \caption{Accuracy as a function of $k$.}
    \label{Accuracy}
\end{figure}



In addition to the above experiments, comparisons were conducted to contrast the performance of our approach when using FastMap and when using Multidimensional Scaling (MDS). Several criteria were considered. As expected, FastMap achieves a faster mapping than MDS. 
It achieves the compression in less than quarter the time needed for MDS on all the considered graphs.

The normalized stress of the mappings was computed for both of the mapping approaches. MDS was able to achieve a lower stress value, which corresponds to a better fitting of the data points, as expected. Yet, the normalized stress values for both approaches seem to converge as $k$ gets larger. As a matter of fact, the difference between such stress values is almost negligible ($\approx$ $10^{-4}$) when $k$ gets larger than 6.
Furthermore, both FastMap and MDS exhibit a decrease in stress value with higher values of $k$, moving towards a more accurate representation. This follows naturally from the fact that we are reducing the amount of compression as we map the graph to a higher dimensional space. 

Surprisingly, the accuracy of our fuzzy method is much higher with the FastMap mapping, as shown in Figure \ref{mdsfast}. This can be attributed to the fact that FastMap assumes the $n$ points (corresponding to graph nodes) are already in a ($n-1$)-dimensional space, which in our case gives a perfect mapping to a Euclidean space, and then reduces the dimension of the $n$ points in such a way that approximates the said (perfect) mapping rather than just find/approximate the intrinsic dimension. 

\begin{figure}[!htb]
\centering
\begin{subfigure}
  \centering
  \begin{tikzpicture}
    \begin{axis}[
        legend style={nodes={scale=0.7, transform shape}},
        xmin=0, xmax=12,
        ymin=0, ymax=110,
        xtick={0,2,4,6,8,10,12},
        ytick={0,20,40,50,60,80,90,100},
        legend pos=south east,
        ymajorgrids=true,
        grid style=dashed,
    ]
    \addplot[
        color=red,
        mark=dot,
        ]
        coordinates {
        (2,7.349)(3,13.66)(4,21.4)(5,18.09)(6,37.88)(7,38.65)(8,42.85)(9,44.56)(10,45.55)(11,45.52)
        };
        \addlegendentry{$FastMap$}
    \addplot[
        color=blue,
        mark=dot,
        ]
        coordinates {
        (2,21.55)(4,21.55)(5,46.72)(6,54.64)(7,58)(8,60.9)(9,62.36)(10,63.86)(11,64.56)
        };
        \addlegendentry{$MDS$}
    \end{axis}
  \end{tikzpicture}
  \caption*{(a) Definite accuracy as a function of $k$.}
\end{subfigure}

\bigskip

\begin{subfigure}
  \centering
  \begin{tikzpicture}
    \begin{axis}[
        legend style={nodes={scale=0.7, transform shape}},
        xmin=0, xmax=12,
        ymin=0, ymax=110,
        xtick={0,2,4,6,8,10,12},
        ytick={0,20,40,50,60,80,90,100},
        legend pos=south east,
        ymajorgrids=true,
        grid style=dashed,
    ]
    \addplot[
        color=red,
        mark=dot,
        ]
        coordinates {
        (2,96.91)(3,96.67)(4,96.33)(5,96.41)(8,93.72)(10,93.49)(11,93.6)
        };
        \addlegendentry{$FastMap$}
    \addplot[
        color=blue,
        mark=dot,
        ]
        coordinates {
        (2,56.78)(3,56.78)(4,56.78)(5,56.8)(6,71.64)(7,80.26)(8,85.87)(9,88.47)(10,89.61)(11,90.35)
        };
        \addlegendentry{$MDS$}
    \end{axis}
  \end{tikzpicture}
  \caption*{(b) Fuzzy accuracy as a function of $k$.}
\end{subfigure}

\caption{Comparison between FastMap and MDS.}
\label{mdsfast}

\end{figure}

Another feature of FastMap that explains the above plausible performance is the fact that FastMap projects the graph nodes in a manner that preserves their clustering structure as much as possible, as mentioned earlier. This behavior has been noted in \cite{Khan}, and might also be gauged from the description of the FastMap algorithm in \cite{faloutsos95fastmap}. 

\subsection{Compression of Edge-Lists}

In some cases one might need to compress (to possibly transfer) the text file containing the text-based graph representation, which is often just the list of edges. 
Thus having a small file size might be of interest, especially if multiple large graphs are to be transferred. In Table \ref{tab:comparisonfilesize} below, we compare the text file size of the initial edge list to the one generated by our FastMap algorithm. In addition, and for further size-reduction, we can apply a double-compression by subsequently using a lossless file compression algorithm to compress both representations further. For the file compression phase, we used 7zip as it is packed with Ubuntu (and other operating systems) and it is known to have one of the better compression ratios. The results show that in most cases, our FastMap-based representation can have a similar size to the initial edge list compressed with 7zip (even smaller in some cases). Moreover, in all cases we have a massive difference in the size of the compressed edge list and the compressed FastMap Representation, the sum of their size combined does not reach 100kB.


\begin{table}[!htb]
\centering
\begin{tabularx}{\columnwidth}{|Y|Y|Y|Y|Y|}
\hline
{\bf Dataset Name} & {\bf Edge List Size} & {\bf FastMap Size} & {\bf Compressed Edge List Size} & {\bf Compressed FastMap Size} \\ \hline
ca-AstroPh & 4.7MB & 507kB & 770kB & 13kB \\ \hline
ca-HepPh & 2.4MB & 327kB & 413kB & 8kB \\ \hline
ca-HepTh & 560kB & 333kB & 114kB & 9kB \\ \hline
ca-GrQc & 306kB & 131kB & 61kB & 4kB \\ \hline
ca-CondMat & \vspace{0.5pt} 2.3MB  & \vspace{0.5pt} 642kB & \vspace{0.5pt} 441kB & \vspace{0.5pt} 17kB \\ \hline
Brightkite & 4.6 MB & 1.1MB & 802kB & 1.4kB \\ \hline
emailEnron & 3.7 MB & 662kB & 578kB & 0.5kB \\ \hline
Catster households & \vspace{0.5pt} 5.9MB & \vspace{0.5pt} 1.2MB & \vspace{0.5pt} 812kB & \vspace{0.5pt} 0.7kB \\ \hline
\end{tabularx}
\caption{Graph representation text file sizes for $k=2$.}
\label{tab:comparisonfilesize}
\end{table}

\subsection{Dealing with Larger Graphs}

Dealing with very large graphs can prohibit the input from being presented in the form of an adjacency (or distance) matrix on a single machine. In fact, the size of the graphs we used in our experiments is at the limit of what can be processed on a single conventional (personal) computer. 
In some applications where the input graph is too large to fit on a single machine, one would expect the input to be distributed across multiple machines. 
In this case, our method can be easily adapted and used by performing dimensionality reduction in a distributed manner as in the work reported in \cite{abu2002} where it was shown that Fastmap can perform well. Our approach would work smoothly in this case since the graph mapping phase is the most space-demanding subroutine. 



\subsection{Discussion}

In general, and despite the difference between our work and the lossless graph compression methods, a simple comparison to the published compression ratios reveals that our technique is far more practical than classical compression methods. Of course, this improvement applies when achieving a high compression ratio is the most crucial measure. In fact, the most used compression techniques seem to target sparse graphs, such as webgraphs. For example, some impressive results were achieved in \cite{boldi2004webgraph}. 
However, unlike our approach, this aforementioned technique exploits two main features of webgraphs when compressing: locality and similarity. Such features are very specific to webgraphs, while our technique is not limited to any special type/class of graphs.
Another common graph compression technique is the one based on Huffman coding 
\cite{Chatterjee2018EfficientDC}, that has also been successfully applied to webgraphs, which are compressed by detecting distinct patterns and replacing them with identifiers that vary in length. This method was shown to yield up to 80\% compression, which is much lower than our achieved ratios. 
Moreover, the said method (used in \cite{Chatterjee2018EfficientDC}) fully exploits the fact that webgraphs are sparse. Such algorithm would perform poorly on dense graphs because many patterns would occur in the adjacency matrix in similar frequency resulting in a low compression ratio. 

As for comparison to lossy methods, we compared our results with the Slim Framework \cite{oltchik}, which contains multiple state of the art methods. The included techniques all output graphs that take up $O(m+n)$ space while our compressed graphs take $O(n)$ space (especially for $k=2$), as mentioned before. When it comes to accuracy, their percentage of critical edges (i.e., ones included in the BFS tree) preserved ranged from $27\%$ to $96\%$. In contrast, we preserved $99.83\%$ of all edges in the worst case.

Finally, despite its fuzzy nature, the overall accuracy of our method makes it close to being lossless. Nevertheless, its most notable feature is the fact that it is ubiquitously effective on all graph types and could reach very high values for large numbers of vertices as well as edges.

\section{Future Directions and Applications}

The approach presented in this paper can be further developed in a number of ways. We hereby list three possible directions:
(i) it can possibly lead to the development of graph algorithms that work well on the compressed graph without having to go back to the original graph, (ii) it can be further enhanced to achieve lossless compression at the cost of loss in compression ratio and (iii) it can be used for network data anonymization.

\subsection{Heuristic Methods}

Consider, for example, the NP-hard Independent Set problem where the objective is to find a largest-possible set of pair-wise non-adjacent nodes. A simple heuristic that works well on Euclidean graphs is to first cluster the graph and then solve the problem locally on each cluster, before collecting all the local solutions.
If our approach is used, we would first adopt a mesh decomposition of the resulting space so that the length of each side of a (square) cell is $R$. This way we guarantee that nodes from non-adjacent cells (even diagonally) are non-adjacent in the original graph. One solution would be to select a maximal set of non-adjacent (non-empty) cells. Another approach would be to solve the problem locally on each cell, or just take a node per cell, and then solve the Vertex Cover problem on the resulting induced subgraph. In both cases we can solve the problem without going back to the original graph.

The above approach can be adopted to solve other graph-theoretic problems, such as Graph Coloring, Dominating Set (using a small cell size) as well as Graph Clustering since the geometric clusters are likely to map back into clusters in the original graphs, as mentioned earlier.


\subsection{Lossless Compression}

For future work on graph compression, we believe our algorithm can be slightly modified in order to achieve lossless compression. Instead of just saving the coordinates of the mapped points, we can also save a list of our "misses" i.e. the edge list consisting of the edges preventing us from having 100\% accuracy. Since our accuracy is already near 100\%, the list will be expected to be small in size and its effect on our compression ratio should be negligible. 

\subsection{Anonymization}

In addition to the potential effective lossless compression, our method can be used as a network anonymization scheme. 
To explicate, after we map a graph into a $d$-dimensional space for some $d$, we can hide the two radii vectors $r$ and $R$, which conceals the adjacency relation completely, making it too difficult (if not impossible) for an adversary to infer any information about the network.
Previous graph anonymization methods focus mainly on the ``hide in the crowd'' approach \cite{degree-anonymization}.
This (additional) anonymization feature of our method is currently under further investigation. Future work considers the objectives of data owners (what to hide and at what cost) and other graph mapping techniques.

\section{Conclusion}

We proposed a lossy method for compressing graphs that makes up for lost information through fuzzy logic. The compression consists of mapping the set of nodes to a set of points in a $k$-dimensional space, via dimensionality reduction, in a manner that attempts at preserving the adjacency relation as much as possible. The distance between a pair of points in addition to another heuristic regarding the number of nodes that lie in a ``fuzzy region'' are to indicate whether or not two nodes were connected by an edge in the original graph. 


Our experiments show a significant percentage of accuracy in the resulting fuzzy representation especially when using FastMap for the dimensionality reduction phase, though it leads to fewer maintained definitive answers when compared to multidimensional scaling (MDS).

Our approach targets large networks, but it would also be interesting to explore the possibility of applying it to small networks in order to enhance modern light-weight computations such as those used in mobile computing platforms \cite{Sharafeddine2, SharafeddineM11}.
In some Machine Learning applications, the compression itself is not as important as achieving some embedding of a graph in a low-dimensional Euclidean space \cite{BarrEmbedding}, which we achieve with very high accuracy.  

The approach presented in this paper was shown to be superior to known graph compression methods when it comes to compression ratio. 
Another potentially plausible aspect of the obtained fuzzy representation stems from our ability to efficiently solve some NP-hard problems approximately by working merely on the compressed graphs, viewed as points in a $k$-dimensional space. 
we believe it would be easy to design heuristic methods for solving problems like Independent Set, Graph Coloring and Clustering (to name a few) without having to go back to the original graph.

It would also be interesting to use the compressed $k$-dimensional representation of a graph to handle $NP$-hard problems on dynamic networks such as the problems discussed in \cite{abuDynamic}. For this, we would need a scheme for possibly partial modification of the Euclidean representation, which can be of interest by itself. 

Finally, further testing can be performed while varying the parameters of the fuzzy inference system described in this paper. As the latter relies on approximate measures, it would be interesting to study whether other membership functions or operators could also be effective.

\bibliographystyle{elsarticle-num} 
\bibliography{ref}

\end{document}